\pdfoutput=1

\documentclass[11pt]{article}

\usepackage[final]{acl}

\usepackage{times}
\usepackage{latexsym}

\usepackage[T1]{fontenc}

\usepackage[utf8]{inputenc}

\usepackage{microtype}

\usepackage{inconsolata}

\usepackage{graphicx}

\usepackage{array}     
\usepackage{longtable} 
\usepackage{float}     

\usepackage{inconsolata}
\usepackage{booktabs}
\usepackage{enumitem}
\usepackage{adjustbox}
\usepackage{subcaption}
\usepackage{comment}

\title{\textsc{Hate}\textcolor{red}{\textsc{PRISM}}: Policies, \textcolor{red}{P}latforms, and \textcolor{red}{R}esearch \textcolor{red}{I}ntegration.\\Advancing NLP for Hate \textcolor{red}{S}peech Proactive \textcolor{red}{M}itigation}

\author{
 \textbf{Naquee Rizwan\textsuperscript{1}},
 \textbf{Seid Muhie Yimam\textsuperscript{2}},
 \textbf{Daryna Dementieva\textsuperscript{3}}, \\
 \textbf{Florian Skupin\textsuperscript{4}},
 \textbf{Tim Fischer\textsuperscript{2}},
 \textbf{Daniil Moskovskiy\textsuperscript{5,6}},
 \textbf{Aarushi Ajay Borkar\textsuperscript{2}}, \\
 \textbf{Robert Geislinger \textsuperscript{2}},
 \textbf{Punyajoy Saha\textsuperscript{1}},
 \textbf{Sarthak Roy\textsuperscript{1}},
 \textbf{Martin Semmann\textsuperscript{2}},\\
 \textbf{Alexander Panchenko\textsuperscript{5,6}},
 \textbf{Chris Biemann\textsuperscript{2}},
 \textbf{and Animesh Mukherjee\textsuperscript{1}}
\\
\\
\small{
 \textsuperscript{1}IIT Kharagpur,
 \textsuperscript{2}University of Hamburg,
 \textsuperscript{3}Technical University of Munich,
 \textsuperscript{4}Bucerius Law School,}\\
\small{
 \textsuperscript{5}Skolkovo Institute of Science and Technology,
 \textsuperscript{6}Artificial Intelligence Research Institute
}
\\
 \small{
   \textbf{Correspondence:} \href{mailto:nrizwan@kgpian.iitkgp.ac.in}{nrizwan@kgpian.iitkgp.ac.in},
   \href{mailto:seid.muhie.yimam@uni-hamburg.de}{seid.muhie.yimam@uni-hamburg.de},
   \href{mailto:daryna.dementieva@tum.de}{daryna.dementieva@tum.de}
 }
}

\begin{document}
\maketitle

\begin{abstract}

Despite regulations imposed by nations and social media platforms, e.g.~\cite{india_it_rules_2021,eu_digital_services_act_2022}, \textit{inter alia}, hateful content persists as a significant challenge. Existing approaches primarily rely on \textit{reactive measures} such as blocking or suspending offensive messages, with emerging strategies focusing on \textit{proactive measurements} like detoxification and counterspeech. In our work, which we call \textbf{\textsc{Hate}\textcolor{red}{\textsc{PRISM}}}, we conduct a comprehensive examination of hate speech regulations and strategies from three perspectives: \textit{country regulations}, \textit{social platform policies}, and \textit{NLP research datasets}. Our findings reveal significant inconsistencies in hate speech definitions and moderation practices across jurisdictions and platforms, alongside a lack of alignment with research efforts. Based on these insights, we suggest ideas and research direction for further exploration of a unified framework for automated hate speech moderation incorporating diverse strategies.~\footnote{\textcolor{red}{Accepted at ACL (Findings), 2025}}

\end{abstract}

\section{Introduction}
AI continues to advance rapidly across various domains, offering diverse applications. Among these, leveraging AI for societal positive impact~\cite{DBLP:journals/corr/abs-2001-01818} is becoming an important direction to explore. Specifically, in the field of NLP~\cite{jin-etal-2021-good}, one of the important societal applications lies in mitigating \textit{digital violence}~\cite{kaye2019speech}.

Digital violence persists as a pressing issue in online social environments, posing tangible risks to users~\cite{Barbieri2019,kara2022research}. It involves using information and communication technologies to hurt, humiliate, disturb, frighten, exclude, and victimize individuals. This often results in increased anxiety, sadness, tension, and a loss of motivation at work~\cite{Torp_Lokkeberg2023-or}. It includes harmful online activities such as abusive behavior, hate speech, toxic speech and offensive language, significantly affecting an individual's professional and social effectiveness and efficiency~\cite{Fahri2022}.

\begin{figure}[t]
    \centering
    \includegraphics[width=1\linewidth]{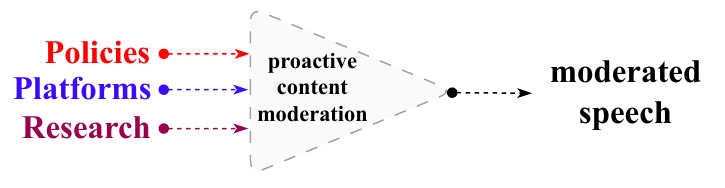}
    \caption{\textbf{\textsc{Hate}\textcolor{red}{\textsc{PRISM}}:} Proactive content moderation with the integration of spectrum involving government and social media platform policies with research.}
    \label{fig:hate_prism}
\end{figure}

\noindent Traditional automated moderation methods typically involve measures such as blocking or suspending accounts that disseminate hateful messages~\cite{macavaney2019hate,cobbe2021algorithmic}. Major technology companies, including Meta and X, have implemented these strategies to manage hate speech. However, such measures have proved insufficient in curbing hateful sentiments over the long term~\cite{parker2023hate}. Alternatives such as counterspeech have gained traction as promising strategies to mitigate hate speech by engaging in dialogue aimed at challenging harmful narratives~\cite{DBLP:journals/corr/abs-2203-03584,kulenovic2023should}. Furthermore, text detoxification represents an approach intended to reduce the toxicity of communications while maintaining the original message~\cite{nogueira-dos-santos-etal-2018-fighting,logacheva-etal-2022-paradetox}. Despite their potential, these approaches have yet to be widely adopted as part of social media platforms' moderation strategies.

\begin{figure}[t]
    \centering
    \vspace{12pt}
    \includegraphics[width=1\linewidth]{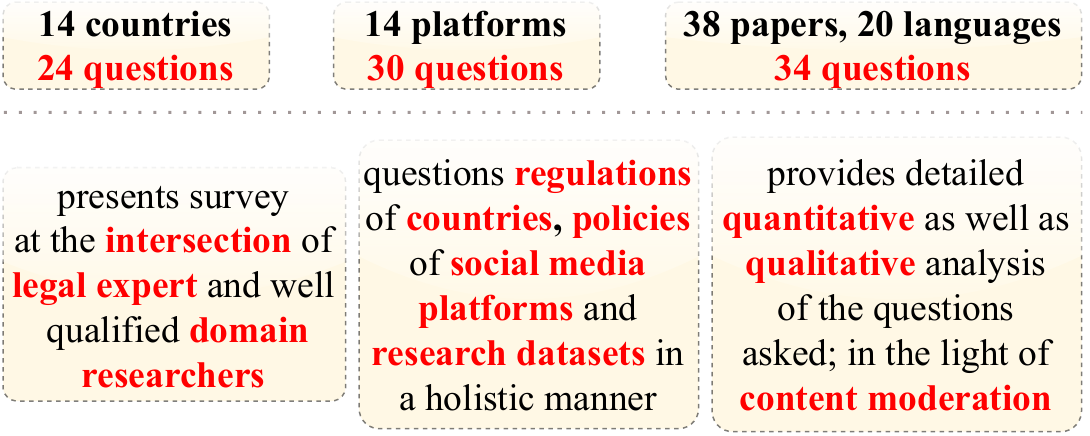}
    \caption{\textbf{Key contributions of this position paper.} \textbf{\textsc{Hate}\textcolor{red}{\textsc{PRISM}}} is the first of its kind that explores hate speech across country-wise regulations, social media platform policies and dataset research papers.}
    \label{fig:summary}
\end{figure}

\paragraph{Key contributions} In this work, \textbf{\textsc{Hate}\textcolor{red}{\textsc{PRISM}}}, we conduct a comprehensive examination of the measures currently employed to mitigate digital violence, focusing on insights drawn from \textit{government regulations}, \textit{social media platform policies}, and \textit{NLP research datasets} (see Figure~\ref{fig:hate_prism}). While our primary objective is to investigate and document these existing frameworks, we also recognize the critical need for empirical evaluation of their practical effectiveness. Our study highlights the current approaches to handle hate speech, emphasizing the disparities and gaps that persist among them. These insights reveal areas ripe for improvement and suggest the need for further empirical research to assess the real-world impact of these measures. Based on our analysis, we propose exploring the potential development of a more unified and cohesive framework in the future to effectively address these gaps. Figure~\ref{fig:summary} presents a brief summary and the survey questionnaire is made available\footnote{\url{https://github.com/hate-alert/platforms\_policies\_research}}. The key contributions of \textbf{\textsc{Hate}\textcolor{red}{\textsc{PRISM}}} are as follows:

\noindent{\textbf{(i)}} We provide a comprehensive survey of hate speech definitions and mitigation strategies from three main perspectives: (a) government regulations across nations; (b) policies of social media platforms; (c) NLP research datasets.\\
\noindent{\textbf{(ii)}} We conduct an extensive comparative analysis of documents from these domains to identify inconsistencies and opportunities for improvement in current moderation practices.\\
\noindent{\textbf{(iii)}} Based on our analysis, we suggest exploring a framework for more formalized methods to combat hate speech in the future.

\paragraph{Key observations} Below we present some of the key brewed insights from our comprehensive study:

\noindent\textbf{(i)} Only \textit{43\%} of the considered countries have regulations of online hate speech and the \textit{USA} is the only country tolerating hate speech; their regulations don't even identify hate speech.\\
\textbf{(ii)} Only \textit{29\%} of the countries have social or community service as punishment and only \textit{21\%} of the nations encourage \textit{proactive content moderation} like counterspeech and text detoxification. This limits hate speech moderation to banning of the users on social media platforms and typically prohibits severe punishments like monetary compensation and/or prison.\\
\textbf{(iii)} Nearly \textit{64\%} of social media platforms encourage counterspeech and text detoxification.\\
\textbf{(iv)} Only \textit{16\%} of considered research dataset papers are aligned with countries’ regulation and just \textit{8\%} align with data sources’ regulations. These observations showcase a wide research gap in the current study of hate speech pointing towards misalignment of dataset papers with countries’ and social media platforms’ regulations.

\section{Related Work}
\paragraph{Digital violence} Violence is an umbrella term that refers to words or actions that cause harm to an individual or a community. Digital violence is a special form that anchors digital technologies, with harm typically spread through electronic devices such as computers, smartphones, and IoT sensors. This form of violence can occur publicly on social media platforms or privately on personal devices and in alternative digital environments like the metaverse.
Our study focuses on digital violence, more specifically, expressed in a textual form. \citet{banko-etal-2020-unified} classified harmful content as either \textit{abusive} or \textit{online harm} and offered a corresponding typology. The typology includes four categories: \textit{hate and harassment}, \textit{self-inflicted harm}, \textit{ideological harm}, and \textit{exploitation}. The study by \citet{lewandowska2023annotation} categorizes harmful content as \textit{offensive speeches}, including 17 sub-categories like \textit{taboo}, \textit{insulting}, \textit{hate speech}, \textit{harassment}, and \textit{toxic}.

\paragraph{Automatic hate speech detection} 
Moderation is a fundamental element of social media platforms, involving various measures to limit the visibility of hateful content. These measures range from deleting and hiding posts to issuing warnings or blocking users who fail to adhere to regulations \cite{DBLP:journals/corr/abs-2312-10269}. In line with these moderation efforts, researchers have also focused on improving automatic detection systems. Significant research efforts have been directed toward gathering datasets that enable the development of automatic hate speech classification models \cite{fortuna-etal-2020-toxic2,mathew2021hatexplain}. These datasets support the creation of models capable of detecting hate speech across various contexts, including those in low-resource languages such as Amharic \cite{ayele-exploring2024}, Arabic \cite{magnossao-de-paula-etal-2022-upv,alzubi-etal-2022-aixplain}, code-mixed Hindi \cite{bohra-etal-2018-dataset,ousidhoum-etal-2019-multilingual}, etc.

\paragraph{Social media platform content policy:} Comparison of various different content moderation strategies can be a time-consuming task because of the diverse formulations and approaches (refer to a recent blog\footnote{\url{https://about.fb.com/news/2025/01/meta-more-speech-fewer-mistakes/}} from Meta). Most often, only single platforms are analyzed by researchers \cite{10.1145/3274301,fiesler2018reddit}. The platforms under study in these works are not chosen in a strategic manner, thus undermining the diverse medium of spread of hate speech. The work by \citet{10.1145/3613904.3642333} proposed an approach for automated collection and the creation of a unified schema to compare platforms. This identified significant structural differences between the platforms in how they deal with these requirements.

\paragraph{Proactive content moderation}
While access restrictions remain a common strategy supported by platforms and government policies to combat harmful content, countering hate speech through engagement is gaining recognition \cite{Mathew_Saha_Tharad_Rajgaria_Singhania_Maity_Goyal_Mukherjee_2019,kulenovic2023should,mun2024counterspeakersperspectivesunveilingbarriers,DBLP:journals/corr/abs-2403-00179, chung-etal-2024-understanding,saha2024zeroshotcounterspeechgenerationllms}. This approach, often encapsulated by the phrase \emph{countering rather than censoring}, is seen as preferable to outright censorship, as it tends to respect the principle of free speech \cite{DBLP:conf/emnlp/YuZ0H23,bonaldi2024nlp}. Beyond reducing hate, \textbf{counterspeech} efforts are utilized to foster positive transformations within online communities by promoting discussions and cultivating a community sense \cite{buerger_why_2022,doi:10.1177/20563051211063843}. 
Another promising avenue in combating toxicity involves text \textbf{detoxification}, which targets eliminating offensive content in messages while preserving the intended meaning \cite{logacheva-etal-2022-paradetox,DBLP:journals/mti/DementievaMLDKS21,tran-etal-2020-towards,dementieva-etal-2025-multilingual,nogueira-dos-santos-etal-2018-fighting}. Detoxification should be viewed as a suggestive tool that recommends less toxic wording, leaving it to the individual user or moderation framework to adopt these changes and enhances the quality of online interactions by facilitating more respectful and less toxic communications \cite{tran-etal-2020-towards}. Various models applied to detoxification\footnote{\url{https://huggingface.co/textdetox}} aim to generate acceptable and diverse non-toxic outputs~\cite{dementieva2024overview}. Contrary to concerns about infringing on freedom of speech, both counterspeech and detoxification contribute to more civil discourse by offering voluntary and non-coercive means of improving online interactions. Both approaches serve as valuable alternatives to traditional moderation methods by promoting positive interaction and personal agency in the moderation process.

\paragraph{Mitigation strategies in deployment} 
\citet{DBLP:journals/osnm/ChungTTG21} developed a tool for Twitter (now X) designed to continuously monitor and respond to hateful content related to Islamophobia. The tool was used by non-governmental organization (NGO) operators, and the counter-narrative feature has been highly praised for its potential to significantly impact the fight against online Islamophobia. Further, \citet{DBLP:journals/csur/AroraNHSNDZDBBA24} in their study examined research on hate speech and related platform moderation policies. The findings reveal a notable discrepancy between the focus of research and the needs of platform policies. This mismatch underscores a gap between the types of content platforms that need to be moderated and the solutions offered by current research on harmful content detection.

\paragraph{Lack of consensus} When it comes to defining \textit{hate speech}, there is no consensus among legislators, platform operators, and researchers~\cite{brown2015hate}. One of the most comprehensive definitions widely followed in the computer science literature, as proposed by the United Nations,\footnote{\url{https://www.un.org/en/hate-speech/understanding-hate-speech/what-is-hate-speech}} describes hate speech as any kind of communication in speech, writing, or behavior that attacks or uses pejorative or discriminatory language with reference to a person or a group based on who they are. To address textual digital violence, traditional automated moderation practice often involves \textit{content moderation} measures. Content moderation, both human and algorithmic, involves overseeing user-generated content to align with legal standards, community norms, and platform policies~\cite{banko-etal-2020-unified,hietanen2023towards}. Algorithmic moderation, primarily aimed at \textit{removing} or \textit{banning} non-compliant content, boosts online safety, curbs abuse, and swiftly detects serious infractions, thus reducing the limitations of depending entirely on human moderators.

\paragraph{Our work} In \textbf{\textsc{Hate}\textcolor{red}{\textsc{PRISM}}}, we identify the gaps between regulatory policies from countries, policies from social media platforms, and approaches used in NLP research. We provide multiple insights and suggest the potential for a more proactive moderation approach to address these challenges.

 
\section{Methodology}
\label{methods}

In \textbf{\textsc{Hate}\textcolor{red}{\textsc{PRISM}}}, our approach incorporates a strategic analysis of hate speech regulation \& mitigation through three primary perspectives: \textit{country-specific regulations}, \textit{social media platforms' policies} and \textit{NLP research approaches}.

\noindent For each of the three perspectives, we developed specific \textit{{Selection Criteria}} to obtain representative samples and crafted a series of \textit{{Questions}} to analyze and gain deeper insights into each area. This dual strategy ensures a comprehensive examination of the regulatory landscape and the effectiveness of various moderation techniques. Furthermore, our analysis also aims to examine three common approaches of content moderation: \textit{blocking/suspending hateful content}, \textit{detoxification of toxic language}, and \textit{counter of hate speech} to engage users constructively. Hence, these moderation techniques were also considered during curation.

\begin{figure} [!ht]
    \centering
    \includegraphics[width=0.85\linewidth]{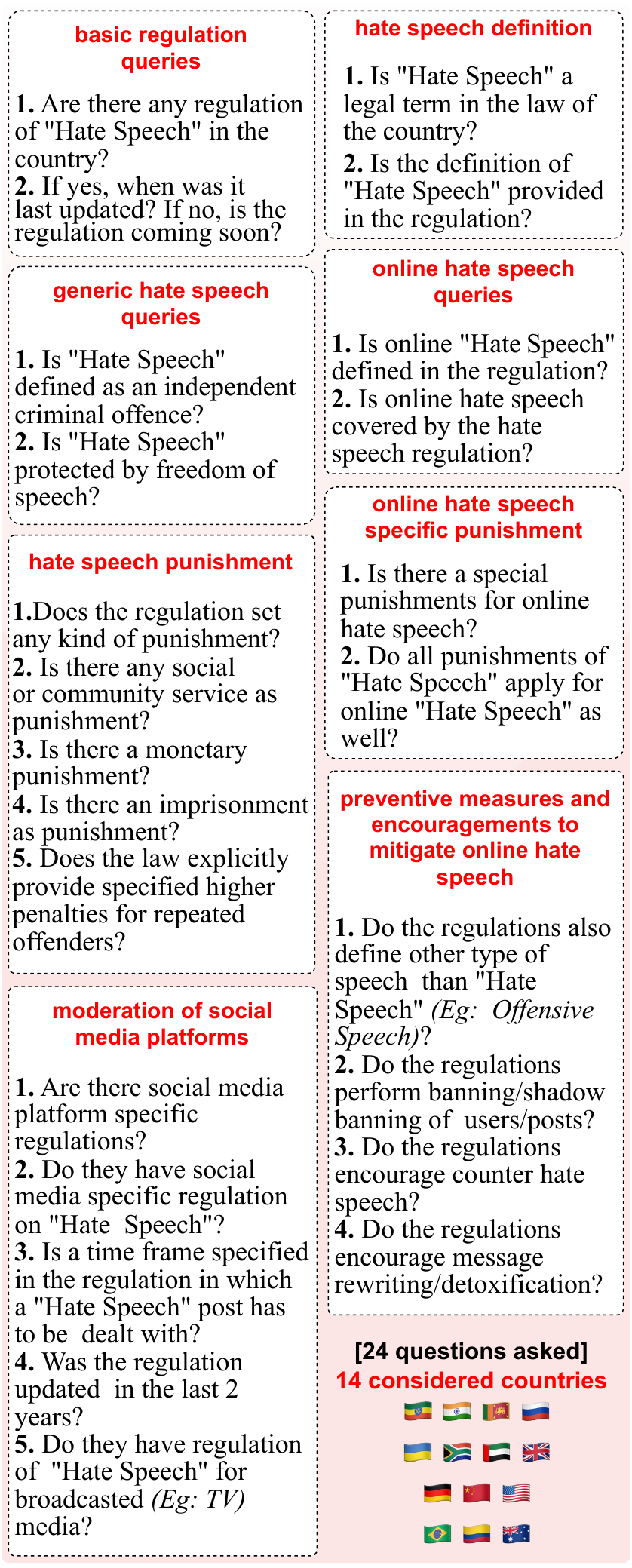}
    \caption{{{\textbf{Country-specific regulations.}} Selected countries and full list of questionnaire encapsulated within categories.}}
    \label{fig:country}
\end{figure}

\begin{figure} [!ht]
    \centering
    \includegraphics[width=0.97\linewidth]{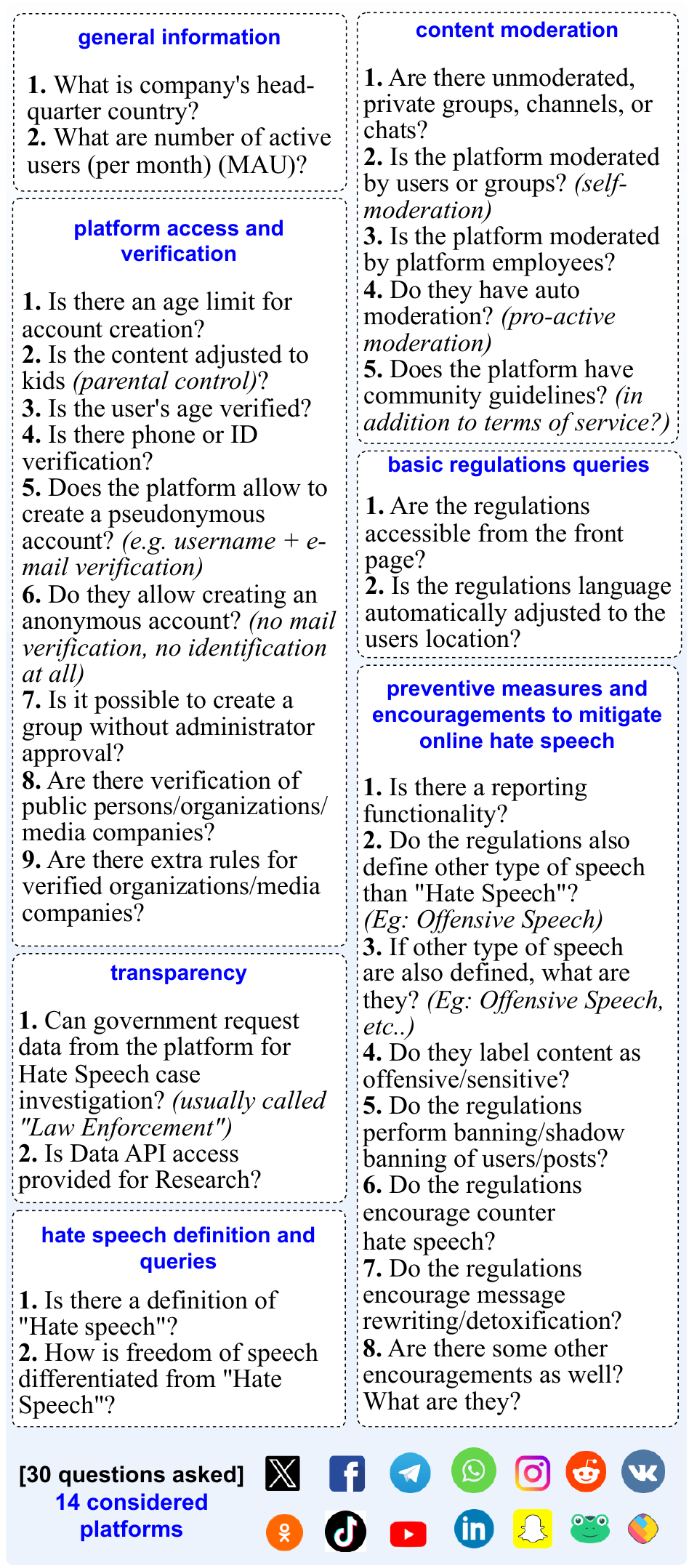}
    \caption{{{\textbf{Platform policies.}} Selected platforms and full list of questionnaire encapsulated within categories.}}
    \label{fig:platforms}
\end{figure}

\paragraph{{Questions}} For each of the three dimensions, we first tunneled down \textit{categories} and then brewed relevant questions for each category (refer Figures~\ref{fig:country},~\ref{fig:platforms},~\ref{fig:research}). These two steps were specifically performed to audit different perspectives in a relevant and robust manner. Note that the surveys used in this study were carefully designed and answered by a group of qualified researchers, including PhD students and postdoctoral fellows, who have expertise in social media policies and online hate speech regulation. Information regarding social media platform policies were gathered through a thorough examination of policy documents and guidelines available on the platforms' official websites.

\paragraph{{Validity assurance}} To ensure the validity and comprehensiveness of our surveys on social media policies and country regulations, we collaborated with \textit{{LegalTech}} expert who an experienced researcher with a Ph.D. and serves as the \textit{executive director} in Legal Technology at Bucerius Law School, with proficiency in IP law, copyright Law, IT law, intellectual property Law, and startup law. This collaboration helped us refine our survey questions and ensure that our research methodologies align with the latest legal and regulatory standards; hence re-assuring that our survey is up to date with the latest regulations of the countries and platforms with their hate speech regulations.

\subsection{Country-specific Regulations}
\label{sec:countries_regulations}
We examine the regulations concerning hate speech that have been established by individual countries. Hate speech can manifest itself in various forms and requires different regulatory approaches depending on cultural, legal, and societal contexts and we maximally incorporate these as discussed below.

\paragraph{{Selection criteria}} To ensure a diverse and representative sample of countries, we selected them based on extensive \textit{familiarity and expertise} of the research team to ensure a detailed and contextually rich analysis. Then we also selected diverse \textit{geographic representation} by selecting at least one country from each continent based on population to capture a wide range of regulatory approaches. Finally, countries with significant \textit{online presence \& engagement} and where incidents of hate speech are prevalent were also considered to strengthen our \textit{focus on hate speech regulation}.

\paragraph{\textbf{{Questions}}} First we narrowed down the categories to have a solid overview of the regulations. For this purpose, we aimed at following rationales for extracting key insights from each country's approach to hate speech regulation. First, we considered \textit{freedom of speech} and \textit{hate speech definition} as they are very crucial for gaining insights into country's tolerance of expression and for reflecting upon their conceptualization and legal stance on hate. Then we considered different \textit{punishments} like monetary fine or imprisonment which is of immense importance; since it is related to the consequences of violation of hate speech regulations. We also employed \textit{preventive measures} to emphasize censorship or content moderation like counterspeech regulatory support and message detoxification. Finally, we also consider \textit{social media regulations} as it is vital for deep diving into regulations related to online hate. After finalizing categories on these key insights, we then added relevant questions into each of them.

\paragraph{\textbf{{Statistics}}} In total, we selected 14 countries from around the world,\footnote{{{Countries:}} Ethiopia, India, Sri Lanka, Russia, Ukraine, South Africa, United States, United Arab Emirates, United Kingdom, Germany, China,  Brazil, Colombia and Australia.} to provide a comprehensive representation of how hate speech and related issues are regulated at the national level. This selection ensures at least one country from each continent is included to capture a diverse set of regulatory approaches and perspectives. Please refer to Figure~\ref{fig:country} for a holistic view. The insights derived from the analysis of these regulations are discussed in Section \ref{regresul}. 

\subsection{Platform Policies}
\label{sec:platform_policies}
We analyze the policies developed by social media platforms to regulate hate speech to understand how the platforms define, detect, and respond to such content. Our analysis provides insights into the accessibility and transparency of platform policies, the use of automated and human moderation, and the preventive measures in place to protect users.

\paragraph{{Selection criteria}} Our selection criteria were designed to ensure a thorough examination of policies across globally popular social media platforms while also accounting for regional variations. \textit{Globally popular platforms} were selected based on their monthly active user count, prioritizing the most widely used platforms worldwide to ensure broad coverage and relevance. For \textit{regionally relevant platforms}, importance was given to the popularity of platforms within the countries mentioned in Section~\ref{sec:countries_regulations}.

\paragraph{{Questions}} We first finalized categories before concluding the final questionnaire. Social media platform specific rationales targeted at distinct aspect of platform functionalities and their strategies for addressing hate speech were considered.
\textit{Hate speech definition} is among the first major rationale we considered for identifying the platform's foundation on content moderation and enforcement actions. Then we pillared on \textit{platform access \& verification}, \textit{regulation accessibility} and \textit{content moderation} as the most crucial rationales. These rationales were chosen to \textbf{(i)} understand the mechanisms for user access and verification, including age restrictions and verification processes, \textbf{(ii)} inquiry into the accessibility and language of platform regulations aimed to assess the transparency, and \textbf{(iii)} help us to further delve into the mechanisms and actors involved in content moderation, including user-driven moderation, automated systems, and employee-led moderation teams. In addition, examination of policy alignment with country-specific regulations provided insights into platform compliance and adaptability to legal frameworks. Similar to rationales in country-specific regulations, here also we include \textit{preventive measures} as they focused on the platform's efforts to empower users in reporting hate speech, as well as initiatives aimed at promoting counterspeech and detoxification of harmful content. Additionally, we include \textit{data access} as an inquiry to assess the platform's transparency and willingness to collaborate with researchers and law enforcement agencies in hate speech investigations.

\paragraph{{Statistics}} 14 social media platforms were selected based on the established selection criteria and analyzed through our detailed questionnaire\footnote{Platforms: X, Facebook, Telegram, WhatsApp, Instagram, Reddit, VK, Odnoklassniki, TikTok, YouTube, LinkedIn, Snapchat, GAB, ShareChat.}.  Figure~\ref{fig:platforms} shows a detailed questionnaire with categories of social media platforms' regulations. The findings, which elucidate the platforms' approaches to hate speech, are presented in Section \ref{platformresult}.

\subsection{Research Datasets}
\label{sec:research_datasets}
\begin{figure} [!ht]
    \centering
    \includegraphics[width=0.97\linewidth]{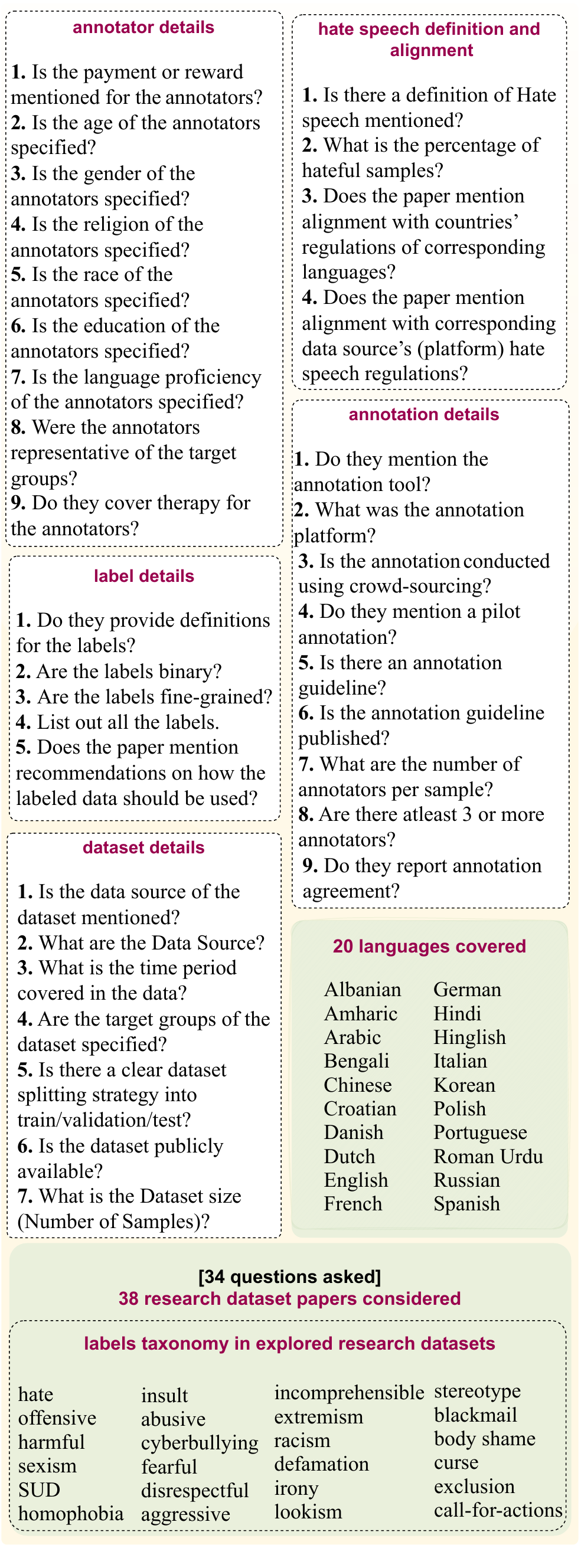}
    \caption{{{\textbf{Research datasets.}} List of covered languages, label taxonomy and full list of questionnaire encapsulated within categories.}}
    \label{fig:research}
\end{figure}

In our third pillar, we bridge the gap with NLP research by examining the current state of automatic \textit{hate speech detection in texts}. Our focus centers on datasets designed for fine-tuning machine learning models, allowing us to gain a comprehensive understanding of the landscape across diverse languages. This exploration will highlight the methodologies used in dataset creation, their definitions of hate speech, and their relevance in addressing the challenges posed by hateful content in digital environments.

\paragraph{{Selection criteria}} Our selection criteria were crafted to ensure the inclusion of diverse perspectives while maintaining a high standard of relevance and credibility. These criteria included the following points --
\textit{Language inclusivity} was chosen as one of the most crucial criterion as it encompasses a wide array of languages prevalent in the countries considered in Section~\ref{sec:countries_regulations}. \textit{Citations} and \textit{publication venue} are one of the most important parameters of success of a research work. We therefore prioritized dataset papers that have significantly influenced the academic community, as indicated by their citation metrics. For low-resource languages, we included the majority or all of the available datasets to ensure comprehensive representation in our analysis. Further, we prioritized \textit{ACL Anthology}, which includes various high-ranked ACL conferences like \textit{ACL}, \textit{NAACL}, \textit{EACL} and \textit{EMNLP}, as well as relevant workshops or venues for shared tasks; specifically the \textit{WOAH} \textit{(Workshop on Online Abuse and Hate)} --constantly co-located with A/A* ACL conferences and indexed in the ACL Anthology-- and other like \textit{SemEval} shared tasks. We also examined the overall proceedings of other relevant conferences like \textit{AAAI}; by paying strict attention to the relevance of papers and their citation metrics.

\noindent Finally, we \textit{cross-verified} the credibility of our selection and choices with established repositories\footnote{\url{https://www.hatespeechdatasets.com}} and refined our paper selection using the highly regarded survey~\cite{Vidgen_2020} on hate speech; thus validating the inclusion of well-established datasets.

\paragraph{{Questions}} For the formulation of categories we designed to extract key insights for a comprehensive understanding of hate speech datasets. \textit{Hate speech definition} is the crucial  here as well as it provides with the complex nature of hate speech, and helps in exploring how researchers conceptualize and define it. Next, we anchor on \textit{annotation process} and diverse set of \textit{labels} as they help in investigating that how the annotation process sheds light on the methodologies employed, including the existence of guidelines, pilot annotations, and quality control measures, which are crucial for evaluating the quality and reliability of the dataset. The labels used for annotation and their descriptions provide insights into the granularity and depth of the dataset's understanding of hate speech nuances. \textit{Annotator demographics} are also very crucial as they help in exploring the demographics of annotators, encompassing factors such as age, gender, religion, and race, facilitated an assessment of dataset inclusivity and annotator suitability. Finally,
\textit{dataset material} which queries aspects such as data source, modality, size, and availability is vital for understanding the dataset's scope and applicability in hate speech research.

\paragraph{Statistics} We selected 38 dataset papers spanning 20 languages based on our criteria and analyzed them using our comprehensive questionnaire. 
The complete questionnaire is available in Figure~\ref{fig:research}, cited datasets are present in Appendix in Table~\ref{tab:datasetCitation} and the results from this analysis are presented in Section~\ref{sec:research_datasets_results}.


\section{Results and Analysis}

In this section, we will discuss the outcomes of our investigation across three key areas aimed at mitigating hate speech: \textit{country regulations}, \textit{platform policies} and \textit{research datasets}. We have summarized our analysis quantitatively in Figures~\ref{fig:country_results},~\ref{fig:platforms_results} and~\ref{fig:research_results} and have also uploaded the full list of questionnaire; as mentioned previously.

\subsection{Regulation Results} 
\label{regresul}

\begin{figure} [!ht]
    \centering
    \includegraphics[width=1\linewidth]{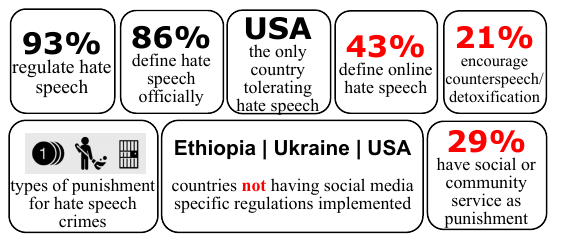}
    \caption{Quantitative results on \textit{country regulations}.}
    \label{fig:country_results}
\end{figure}

As stated earlier, we selected 14 countries from all over the world in order to have a comprehensive picture of how hate speech and related issues are regulated on a governmental level. The quantitative results of our investigation are summarized in Figure~\ref{fig:country_results} and below we perform qualitative analysis.

\paragraph{Key observations} First of all, we note that all countries considered in the study regulate hate speech except the \textit{USA} and the majority of the regulations have been updated no earlier than \textit{four years} ago, keeping the nations up-to-date with the current hate speech challenges. The \textit{definition of hate speech}, inspite of the widespread recognition of the need to address hate speech at the governmental level, lacks single universally accepted definition of what constitutes hate speech. Different countries have developed their own definitions, reflecting their unique cultural, legal, and social contexts. Understanding these context-specific definitions is crucial for developing targeted interventions that respect local norms while safeguarding individuals from harmful speech.

\paragraph{Online hate speech} Although most countries have laws regulating hate speech, only \textit{43\%} have specific definitions related to \textit{online hate speech}. Countries such as the USA, Russia, and Ukraine do not independently address online hate speech at the legislative level, whereas hate speech is protected under freedom of speech in the USA.

\paragraph{Punishments} Most countries adopt various approaches to punish hate speech offenders, with penalties ranging from fines and community service to imprisonment. While imprisonment is a potential consequence, the duration of sentences is typically relatively short, and varies from one country to the other.

\paragraph{Proactive content moderation} Proactive mitigation of hate speech is being used in a limited manner. At both national and regional levels, specific laws addressing counterspeech and detoxification are \textit{lacking}. However, many countries have emphasized the creation of a safe environment through proactive methods, which appears to be a positive initial step in this direction.

\subsection{Platform Results} 
\label{platformresult}
\begin{figure} [!ht]
    \centering
    \includegraphics[width=1\linewidth]{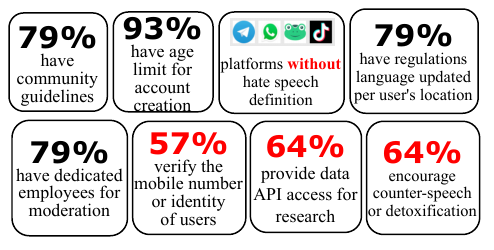}
    \caption{Quantitative results on \textit{platform policies.}}
    \label{fig:platforms_results}
\end{figure}
In this subsection we analyze the outcome of our survey on social media platform's policies to robustly corroborate the community guidelines provided by the respective platforms in terms of hateful content and their mitigation strategies. The overall quantitative results from our investigation are summarized in Figure~\ref{fig:platforms_results}; below, we perform qualitative analysis.

\paragraph{Key observations} The majority of platforms have an age limit for account creation and some sort of parental control. Only \textit{three} out of 14 platforms we studied---Facebook, Instagram, and YouTube---apply age verification methods. Phone number or any other sort of ID verification is present in only \textit{57\%} of the platforms that we studied. None of the platforms allow for the creation of completely \textit{anonymous accounts}, but nine platforms allow for the creation of \textit{pseudonymous accounts}, i.e., an account that uses a fictitious name or alias to protect the user's digital identity.

\paragraph{Community guidelines} All platforms except GAB have made their regulations accessible from their home pages. X, Telegram and GAB are the platforms that \textit{do not adjust the language} of the regulations automatically according to the user's geographical location. Platforms like---Telegram, WhatsApp, TikTok and GAB---do not even have a strict definition of hate speech in their regulations.

\paragraph{Content moderation} Platforms play an important role in content moderation, where administrators or moderators can moderate respective groups or communities. It is highly subjective and dependent on the social and cultural context of the individual and their demographics. Only a small minority of platforms---Facebook, Instagram, TikTok, ShareChat and YouTube---have moderators with \textit{demographic diversity}. A common solution to this challenge is employing \textit{auto-moderation}, which is adopted by almost all platforms except-- Telegram, WhatsApp, and GAB.

\paragraph{Preventive measures} All platforms have a reporting functionality where users can report content they find inappropriate. The users generally flag the reported content according to the category labels provided by the platform. Platforms like---WhatsApp, VK, Odnoklassniki, TikTok, and ShareChat---do not provide a label for hateful or sensitive content when reporting.

\paragraph{Proactive content moderation} At last, we analyze the acceptance of counterspeech and message detoxification as a proactive moderation strategy. Surprisingly, we found very few platforms like---Facebook, VK and Odnoklassniki---that encourage the promotion of these new moderation paradigms.

\subsection{Results based on Research Datasets}
\label{sec:research_datasets_results}

\begin{figure} [!ht]
    \centering
    \includegraphics[width=1\linewidth]{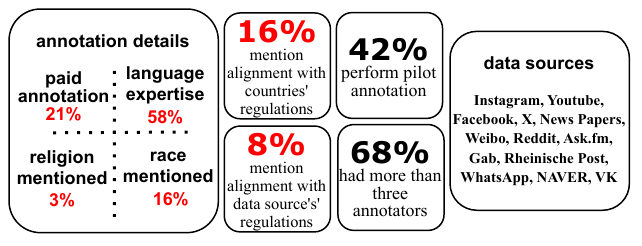}
    \caption{Quantitative results on \textit{research datasets.}}
    \label{fig:research_results}
\end{figure}

Our analysis of various hate speech dataset papers has yielded several key findings that provide insights into the landscape of hate speech research and dataset construction. Quantitative results are provided in \textbf{\textit{Figure~\ref{fig:research_results}}} and below we share qualitative analysis.

\paragraph{Key observations} Interestingly, \textit{66\%} of the surveyed papers present a clear \textit{definition of hate speech} within their work. We believe, especially for annotation tasks and dataset papers, conceptual clarity in understanding hate speech is highly important. Consequently, our expectation was that almost all papers would have a definition of hate speech, which is unfortunately not true. Further, our analysis reveals that only \textit{16\%} of the papers have \textit{cross-checked} their definition with hate speech regulations at the national level, and \textit{only three papers} referenced platform-specific regulations. This lack of alignment with regulatory frameworks highlights potential discrepancies between academic definitions and legal or platform-specific interpretations of hate speech. To our surprise, \textit{only one} of the 38 surveyed papers formulate recommendations on \textit{leveraging their work}, datasets, or annotations. This highlights a missed opportunity for academic research to inform practical interventions and policy-making efforts in the fight against hate speech.

\paragraph{Platform imbalance} Finally, we observe considerable imbalance in investigated \textit{data sources}. X account for over \textit{50\% }of the studies, while other platforms such as---YouTube, Instagram, Reddit and WhatsApp---were explored in less than \textit{10\%} of the papers. Facebook, with its 3 billion users, far exceeds X, which has only 611 million users, indicating that the over-representation of certain platforms does not correlate with actual usage (cf. further insights in Appenix~\ref{appendix:analysis_continued}).

\section{Conclusion and Future Directions}

The challenges identified by \textbf{\textsc{Hate}\textcolor{red}{\textsc{PRISM}}} in addressing hate speech from governmental, platform, and research angles are as following.

\textit{{Firstly}}, the lack of a universally accepted definition of hate speech complicated the development of consistent regulations across countries. While most nations have a definition of hate speech, only a third defined it specifically for the online environment. This underscored the need to raise awareness about online hate and its mitigation. However, some of the countries are proactive in hate speech moderation methods development.

\textit{{Secondly}}, social media platforms showed policy inconsistencies, which hindered effective content moderation. A fifth of the platforms failed to adapt hate definitions to local languages and cultures, and moderation typically focused on banning rather than proactive strategies.

\textit{{Thirdly}}, most NLP research did not align with platform or regulatory guidelines, often reusing outdated definitions from previous computer science studies. Moreover, many studies did not explore proactive measures such as counterspeech or detoxification in operational settings. Such data labeling could improve online hate mitigation.

\textit{{Ultimately}}, collaboration between platforms, governments, and researchers is essential to create dynamic moderation frameworks. Aligning definitions and promoting proactive strategies will lead to more effective solutions for combating online hate. The further exploration of proactive moderation pipeline which consists of thoughtful combination of text detoxification, counter speech generation, other preventing measures, and preparing such datasets for automatic methods development should be a frontier for future research.

\section*{Acknowledgments}
This work was made possible with the high collaboration efforts of the team across nations. All authors would like to thank SPARC-II (Scheme for Promotion of Academic and Research Collaboration, Phase II) project for funding international travel and subsistence to carry out this work. Daryna Dementieva's work was additionally supported by Friedrich Schiedel TUM Fellowship.

\section*{Limitations}

While we made diligent efforts to meticulously document our research process, findings and recommendations, it is important to acknowledge that our study has certain limitations:

\noindent\textbf{1) Only text-based content}: We only took into consideration textual expression of digital violence in NLP research. We acknowledge that hate can also be extremely taxing in other modalities like images, voice recordings and videos. Our study on hate mitigation do not encompass such cases.

\noindent\textbf{2) Only human-written content}: Our mitigation pipeline was initially tailored to address only human-authored messages and comments. However, as text generation systems become more prevalent, there is a growing influx of machine-generated content on social media platforms. It is imperative to incorporate additional measures to detect and address bots and other machine-generated texts that may pose greater risks in inciting hatred.

\noindent\textbf{3) Only digital content}: Finally, we performed our studies only in the realm of digital violence. Nevertheless, digital hater can transcend virtual platforms and manifest in real-world scenarios through various means. For this reason, we include an `authorities' intervention' step in our demarcation pipeline. 

\section*{Ethics statement}
We are committed to upholding freedom of speech and respect the autonomy of stakeholders in deploying moderation technologies tailored to their specific domain, context, and requirements. Our aim is to offer a broader perspective on potential automatic proactive moderation strategies, providing novel insights and recommendations.

\bibliography{custom}

\appendix

\begin{table*}[!ht]
\centering
\footnotesize
\begin{tabular}{c|c}
\toprule
\textbf{Year of publication} & \textbf{Dataset research papers}                                                                                                                                    \\ \midrule
2017 \textit{(2)}                 & \begin{tabular}[c]{@{}c@{}}~\cite{Davidson_Warmsley_Macy_Weber_2017, Fabio_Del_Vigna_2017}\end{tabular}       \\ \midrule
2018 \textit{(7)}                & \begin{tabular}[c]{@{}c@{}}~\cite{8508247, Founta_Djouvas_Chatzakou_Leontiadis_Blackburn_Stringhini_Vakali_Sirivianos_Kourtellis_2018, bohra-etal-2018-dataset, mathur-etal-2018-detecting}\\ ~\cite{sanguinetti-etal-2018-italian, sprugnoli-etal-2018-creating, MEX-A3T-Carmona-2018}\end{tabular}                                                                                                                                       \\ \midrule
2019 \textit{(9)}                & \begin{tabular}[c]{@{}c@{}}~\cite{mulki-etal-2019-l, 10.1007/978-3-030-32959-4_18, chiril-etal-2019-multilingual}\\ ~\cite{ousidhoum-etal-2019-multilingual, 10.1145/3368567.3368584, corazza:hal-02381152}\\ ~\cite{Ptaszynski_Pieciukiewicz_Dybała_2019, fortuna-etal-2019-hierarchically, basile-etal-2019-semeval}\end{tabular}                       \\ \midrule
2020 \textit{(5)}                & \begin{tabular}[c]{@{}c@{}}~\cite{MOSSIE2020102087, sigurbergsson-derczynski-2020-offensive}\\ ~\cite{bhardwaj2020hostility, rizwan-etal-2020-hate, zueva-etal-2020-reducing}\end{tabular}                       \\ \midrule
2021 \textit{(6)}                & \begin{tabular}[c]{@{}c@{}}~\cite{9564230, 10.1007/978-981-16-0586-4_37, 11356/1462}\\ ~\cite{burtenshaw-kestemont-2021-dutch, mathew2021hatexplain, NEURIPS_DATASETS_AND_BENCHMARKS2021_c9e1074f}\end{tabular}                       \\ \midrule
2022 \textit{(8)}                & \begin{tabular}[c]{@{}c@{}}~\cite{nurce2022detecting, 10.1007/978-3-030-93709-6_41, 9971189, jeong2022kold}\\ ~\cite{das-etal-2022-hate-speech, JIANG2022100182, shekhar-etal-2022-coral, demus-etal-2022-comprehensive}\end{tabular}                       \\ \midrule
2023 \textit{(1)}                & \begin{tabular}[c]{@{}c@{}}~\cite{10076443}\end{tabular} \\ \bottomrule
\end{tabular}
\caption{{\textbf{{Explored Research Datasets:}} Arranged in ascending chronological order. Number in brackets denote the number of explored dataset papers published in the corresponding year.}}
\label{tab:datasetCitation}
\end{table*}

\section{Further insights}
\label{appendix:analysis_continued}
Deep investigation into hate speech dataset papers revealed a nuanced understanding of hate speech as a multi-faceted phenomenon. Through the analysis of hate speech definitions and descriptions, several key aspects emerged that can be considered for the classification of hate speech. We outline these aspects below:\\

\noindent\textbf{(i) Target:} Understanding the target of hate speech is essential in contextualizing its impact. Inflammatory messages directed at individuals or groups are often considered hate speech, while undirected messages are not.\\
\textbf{(ii) Discrimination:} Hate speech often manifests through discriminatory language targeting various characteristics such as race, sex, gender, nationality, religion, and more.\\
\textbf{(iii) Intent of the perpetrator:} Malicious intent, ranging from mocking and causing emotional harm to issuing threats or inciting violence, is typical for hate speech. However, humorous, sarcastic, or troll messages are often not considered hate speech.\\
\textbf{(iv) Language usage}: Hate speech can manifest in diverse linguistic forms, from threatening, dehumanizing, or fear-inducing speech to overtly violent or obscene language. Again, sarcastic or humorous language is often not considered hate speech.\\
\textbf{(v) Emotions of the victim/target}: Understanding the emotional impact on hate speech victims is crucial for assessing its harm, as it often induces sadness, anger, fear, and out-group prejudice.\\
\textbf{(vi) Frequency}: Hate speech can manifest as isolated incidents or persistent harassment, such as mobbing or bullying. Analyzing attack frequency helps gauge the severity of hate speech.\\
\textbf{(vii) Time}: Hate speech may reference past events, current circumstances, or future actions. Especially, messages that incite violent actions in the near future are dangerous. The temporal dimension should not be neglected.\\
\textbf{(viii) Fact-checking}: Hate speech often relies on misinformation or distorted facts to perpetuate harmful narratives. Identifying disinformation can aid hate speech detection and inform the severity.\\
\textbf{(ix) Topic and context}: Hate speech targets various topics, from political ideologies to social identities, and contextual factors must be considered in its assessment.
Our analysis underscores the complexity of hate speech, highlighting the need for nuanced approaches to effectively identify, classify, and mitigate its harmful effects.

\end{document}